\documentclass{article}

\usepackage{arxiv}

\usepackage[utf8]{inputenc} 
\usepackage[T1]{fontenc}    
\usepackage{hyperref}       
\usepackage{url}            
\usepackage{booktabs}       
\usepackage{amsfonts}       
\usepackage{nicefrac}       
\usepackage{microtype}      
\usepackage{lipsum}
\usepackage{graphicx}
\graphicspath{ {./images/} }

\title{Positional Masking for Language Models}

\author{
 Andy Wagner \\
  Microsoft\\
  \texttt{anwagner@microsoft.com} \\
  \And
  Tiyasa Mitra \\
  Microsoft \\
  \texttt{Tiyasa.Mitra@microsoft.com} \\
  \AND
  Mrinal Iyer \\
  Graphcore \\
  \texttt{mrinali@graphcore.ai} \\
  \And
  Godfrey De Costa \\
  Graphcore \\
  \texttt{godfrey.da.costa@graphcore.ai} \\
  \And
  Marc Tremblay \\
  Microsoft \\
  \texttt{marc.tremblay@microsoft.com} \\
}

\title{Positional Masking for Language Models}

\author{%
  Andy Wagner \\
  Microsoft\\
  \texttt{anwagner@microsoft.com} \\
  \And
  Tiyasa Mitra \\
  Microsoft \\
  \texttt{Tiyasa.Mitra@microsoft.com} \\
  \AND
  Mrinal Iyer \\
  Graphcore \\
  \texttt{mrinali@graphcore.ai} \\
  \And
  Godfrey De Costa \\
  Graphcore \\
  \texttt{godfrey.da.costa@graphcore.ai} \\
  \And
  Marc Tremblay \\
  Microsoft \\
  \texttt{marc.tremblay@microsoft.com} \\
}

\begin{document}

\maketitle

\begin{abstract}
  Masked language modeling (MLM) pre-training models such as BERT\cite{devlin-etal-2019-bert} corrupt the input by replacing some tokens with [MASK] and then train a model to reconstruct the original tokens. This is an effective technique which has led to good results on all NLP benchmarks. We propose to expand upon this idea by masking the positions of some tokens along with the masked input token ids. We follow the same standard approach as BERT\cite{devlin-etal-2019-bert} masking a percentage of the tokens positions and then predicting their original values using an additional fully connected classifier stage. This approach has shown good performance gains (.3\% improvement) for the SQUAD\cite{rajpurkar-etal-2016-squad} task in general along with an additional improvement in convergence times. For the Graphcore IPU\cite{Graphcore:2020} the convergence of BERT Base with position masking requires only 50\% of the tokens from the original BERT paper. 
\end{abstract}

\section{Introduction}
Self training methods based on models using Transformer\cite{Vaswani2017AttentionIA} blocks like BERT\cite{devlin-etal-2019-bert} and it's descendants like XlNet\cite{Yang2019XLNetGA}, Albert\cite{Lan2020ALBERTAL}, Roberta\cite{Liu2019RoBERTaAR} and many others have brought significant performance gains for NLP tasks. We are enhancing the training approach of these models by masking the positions along with the token ids. Our simulations are focused on the BERT architecture but the results should scale to other networks as well given we are just supplying extra information to the network.

In Bert a small subset of the unlabeled input sequence is masked or replaced and the network is trained to recover the input. We enhanced this approach by performing a similar masking operation on the token positions. The technique has a few advantages. First it gives the network extra information to train leading to quicker convergence and greater stability. Second it helps by adding extra information in training which is normally limited by masking a small percentage of tokens. Third it improves the training of the position encodings which have a large impact on the performance.

\section{Improvements}
The focus of this paper is position masking but we also found that by allowing all gradients to flow through the dropout layer led to substantially better squad results (~.5\%). 

 \begin{figure}
  \centering
 
  \rule[-.5cm]{0cm}{4cm} 
  \includegraphics[width=0.7\columnwidth]{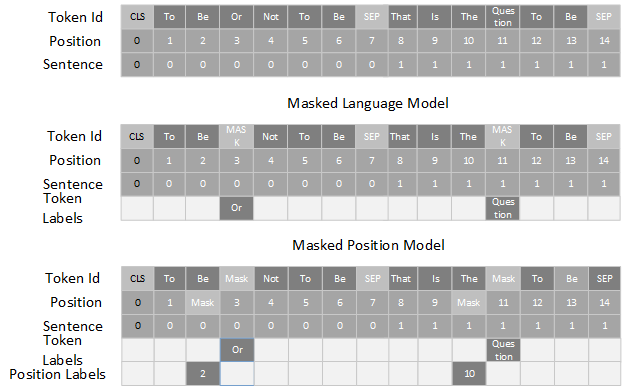}
    \label{fig:fig1}
  \rule[-.5cm]{4cm}{0cm}
  \caption{Example of position masking}
\end{figure}

\begin{figure}
  \centering
 
  \rule[-.5cm]{0cm}{4cm} 
  \includegraphics[width=0.7\columnwidth]{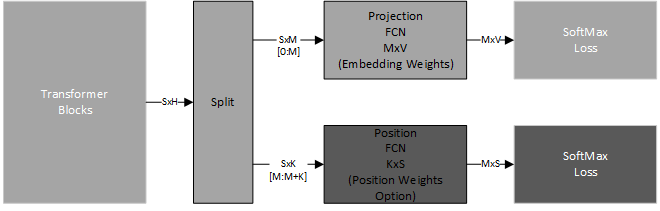}
   \label{fig:fig2}
  \rule[-.5cm]{4cm}{0cm}
\caption{Diagram of changes from bert architecture}

\end{figure}

\subsection{Position Masking}

Position masking requires a single extra classifier layer to be added to a transformer based system. A standard BERT implementation with an extra fully connected classifier stage targeted towards estimating the location of the masked position tokens is shown in figure \ref{fig:fig2} with the changes in a darker background. This diagram assumes that the tokens are packed in the embedding layer to allow an efficient implementation at the classifier. A more standard gather/scatter operation could also be used. 

The concept behind position masking is a somewhat obvious extension intuitively although leads to a slightly different analytic solution than masking the token ids. The position masking problem optimizes the network to solve the location of the position encoding which differs from the desired solution of identifying the masked token. While this leads to some inefficiency in the network, this effect is overcome due to the tighter control over the position encodings and greater information accessible to the network. We have not considered approaches where finding the position of a masked token could be used directly or used in a different way to enhance networks using a less direct approach but believe there are approaches possible.

\subsection{Enhanced Dropout Gradient}
For fine tuning tasks, although we dropout the attention weights in the forward, we do not apply the dropout mask when the gradients are back-propagated. We have seen that for some tasks this gives a consistently better performance over multiple runs than applying the dropout mask as would typically be expected when dropout is used.

\section{Experimental Approach}
 We chose BERT\cite{devlin-etal-2019-bert} Base for to use as the baseline for out study with Squad\cite{rajpurkar-etal-2016-squad} as the performance metric. We chose two implementations to study for comparison purposes GPUs and IPUs. The majority of our work has been done on IPUs due to it's performance advantage but GPU results were included given it is a more mature technology.  
 
 Our approach consists of masking the position encodings along with the masked token ids. A simple example of this is shown in figure \ref{fig:fig1}. We use the same masking strategy for position as BERT\cite{devlin-etal-2019-bert} used for token ids. 90\% of the tokens were masked with 5\% usign the correct position and 5\% using a random token. 
 
 We have for the most part stayed true to the original approaches from the BERT paper \cite{devlin-etal-2019-bert}. We used Wikipedia as a training set and tokenized the data using the original BERT code base. We loosely followed the standard approach which consists of 90\% training with sequence length 128 followed by 10\% of the training using sequence length 384. We did not necessarily stay true to the split times and optimized for efficiency. For Base, we chose to have a more even split on IPUs due to the scaling efficiency for that case. We chose not to use BookCorpus as early experiments didn't show an advantage. 
 
  We chose a pipelined implementation of BERT using Graphcore IPU which was standard from an algorithmic perspective with the exception that we chose to use standard SGD rather than ADAM as well as adding a new technique in fine tuning in handling the gradient through the dropout layers. We also used the Nvidia Mixed Precision implementation which used the LAMB optimizer with default settings supplied for comparison with a more mature technology.

\section{GPU Results}
\begin{figure}
  \centering
  \rule[-.5cm]{0cm}{4cm} 
    \begin{minipage}[t]{0.4\linewidth}
    \centering
    \includegraphics[width=1.0\columnwidth]{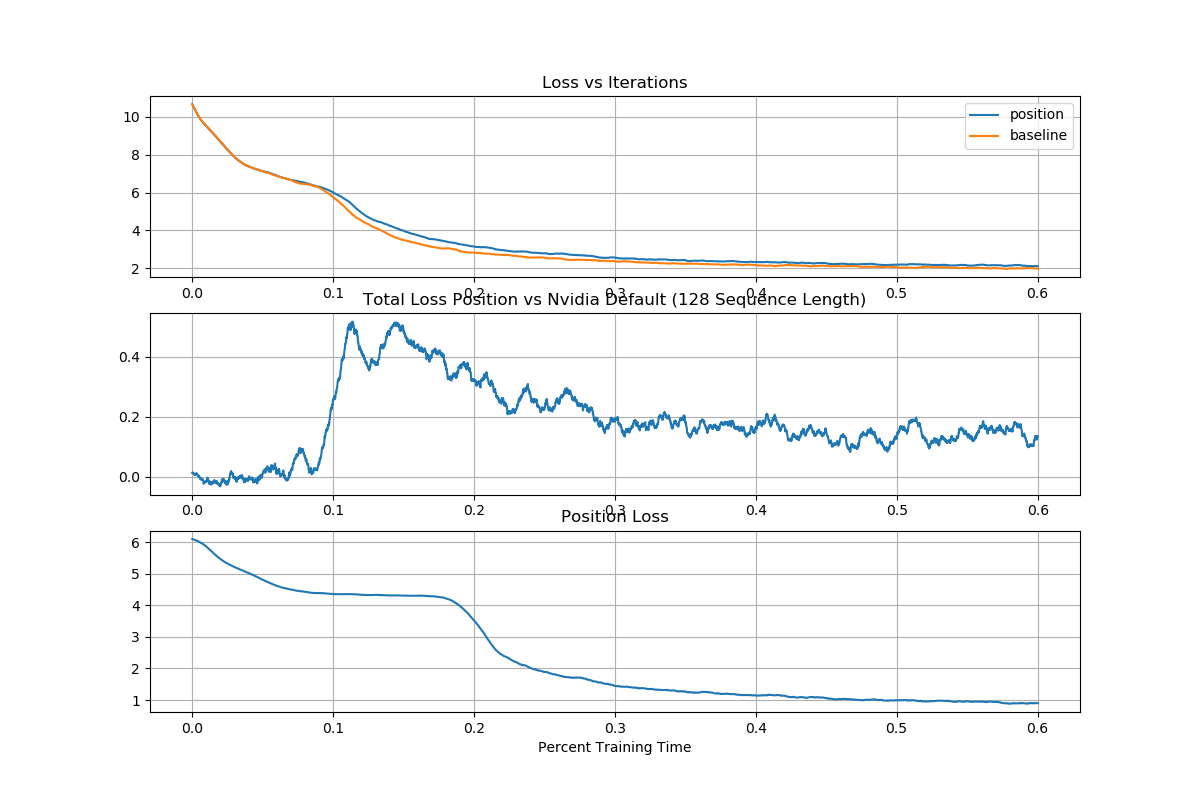}
    \caption{GPU phase 1 MLM accuracy} \label{fig1:side:a}
    \end{minipage}%
    \hspace{1cm}%
    \begin{minipage}[t]{0.4\linewidth}
    \centering
    \includegraphics[width=1.0\columnwidth]{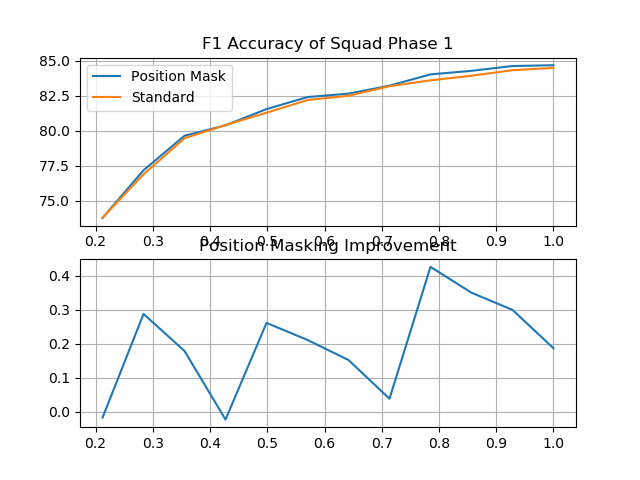}
    \caption{GPU phase 1 Squad Accuracy} \label{fig1:side:b}
    \end{minipage}
  \rule[-.5cm]{4cm}{0cm}
\end{figure}

The GPU results were created using the PyTorch version of Nvidia's mixed precision directly from their website\cite{Nvidia:2020} with the default settings. The only changes made was to use a 384 sequence length rather than 511 to allow better comparisons with the IPU results. For Phase 1 tracking the GPU performance was better after the warmup phase by .3\% which carried over to phase 2. The results for phase 1 MLM and Position loss are shown in figure \ref{fig1:side:a} and the squad performance for this data is shown in figure \ref{fig1:side:b}. The performance with position masking is about .3\% throughout the run with the exception of the warm-up period. 

The results from phase 2 are shown in table \ref{tab:table1}. The GPU performance was .3\% better at the end of the run. These results were averaged over multiple pre-training and squad runs but never reached the published BERT Base results. This is partially due to using sequence length 384 which degrades performance and partially due to not looking for the peak performance. The results in the paper are all based on averages to better compare techniques.

\begin{table}
 \caption{GPU phase 2 bert base performance}
  \centering
  \begin{tabular}{lll}
    \toprule
    \cmidrule(r){1-2}
    Name     & Squad v1.1 F1       \\
    \midrule
    Base 384 &   87.99     \\
    Position 384 & 88.26       \\
    \bottomrule
  \end{tabular}
  \label{tab:table1}
\end{table}

\section{IPU Results}
The IPU has a general advantage over the GPU for BERT Base on the Squad Task. It converges in about 60\% of the tokens of the original paper with 1\% better performance results. The addition of position masking added .3\% F1 performance improvement and .4\% EM performance improvement on top of the results and brings the convergence time down to 50\% of the tokens used in the original paper. The difference in EM performance is probably due to the tighter control over positions. 

\begin{figure}
  \centering
  \rule[-.5cm]{0cm}{4cm} 
    \begin{minipage}[t]{0.4\linewidth}
    \centering
    \includegraphics[width=1.0\columnwidth]{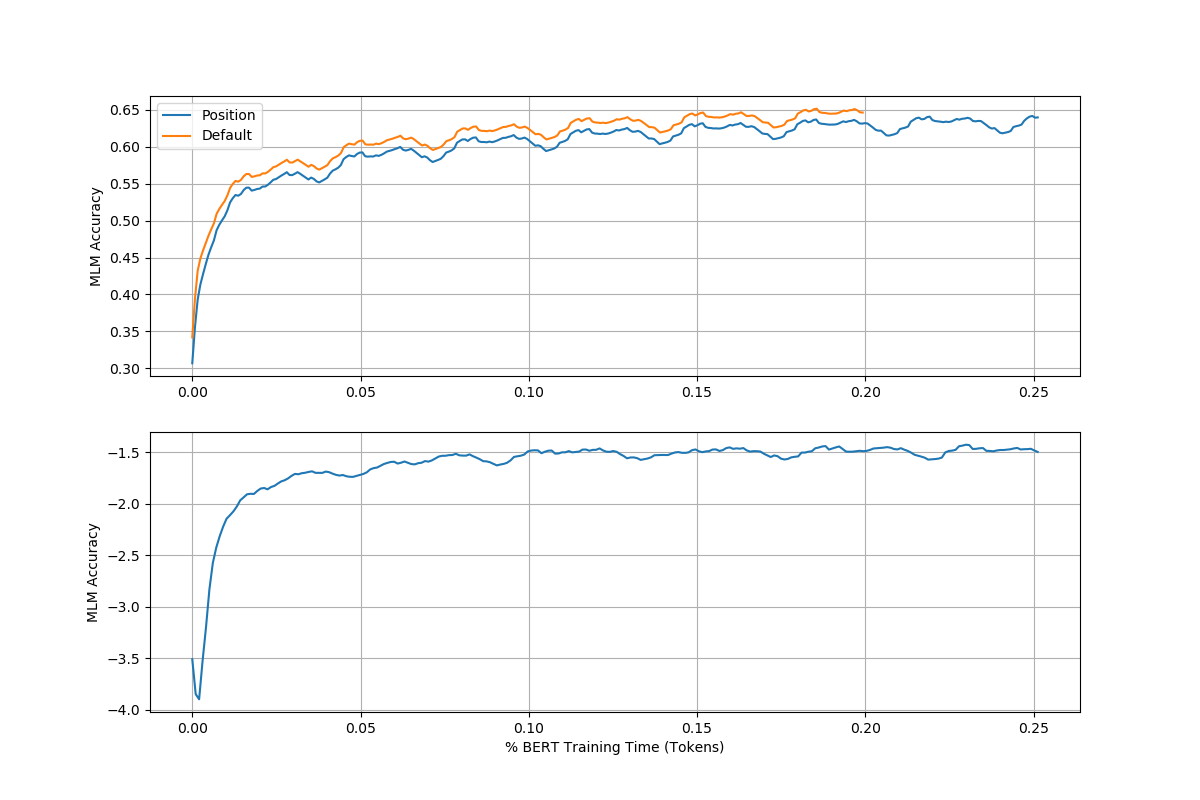}
    \caption{IPU phase 1 MLM accuracy} \label{fig2:side:a}
    \end{minipage}%
    \hspace{1cm}%
    \begin{minipage}[t]{0.4\linewidth}
    \centering
    \includegraphics[width=1.0\columnwidth]{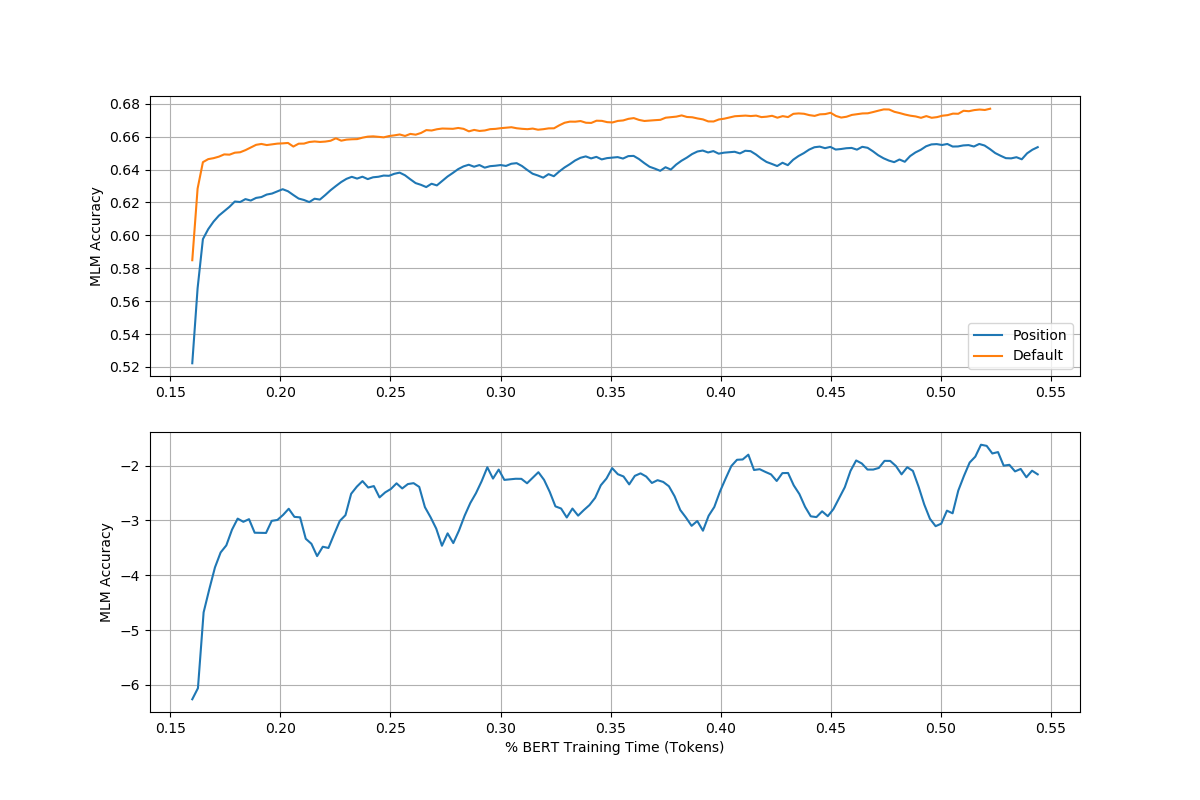}
    \caption{IPU phase 2 MLM accuracy} \label{fig2:side:b}
    \end{minipage}
  \rule[-.5cm]{4cm}{0cm}
\end{figure}

The results for phase 1 MLM convergence with position masking is shown compared to the baseline mode in figure \ref{fig2:side:a}. The baseline mode has a 1.5\% better MLM performance which is due to the masking of positions resulting in lost information which degrades the MLM performance. The phase 2 results are shown in figure \ref{fig2:side:b} and has a similar response where the performance is 2\% lower. 

\begin{figure}
  \centering
  \rule[-.5cm]{0cm}{4cm} 
    \begin{minipage}[t]{0.4\linewidth}
    \centering
    \includegraphics[width=1.0\columnwidth]{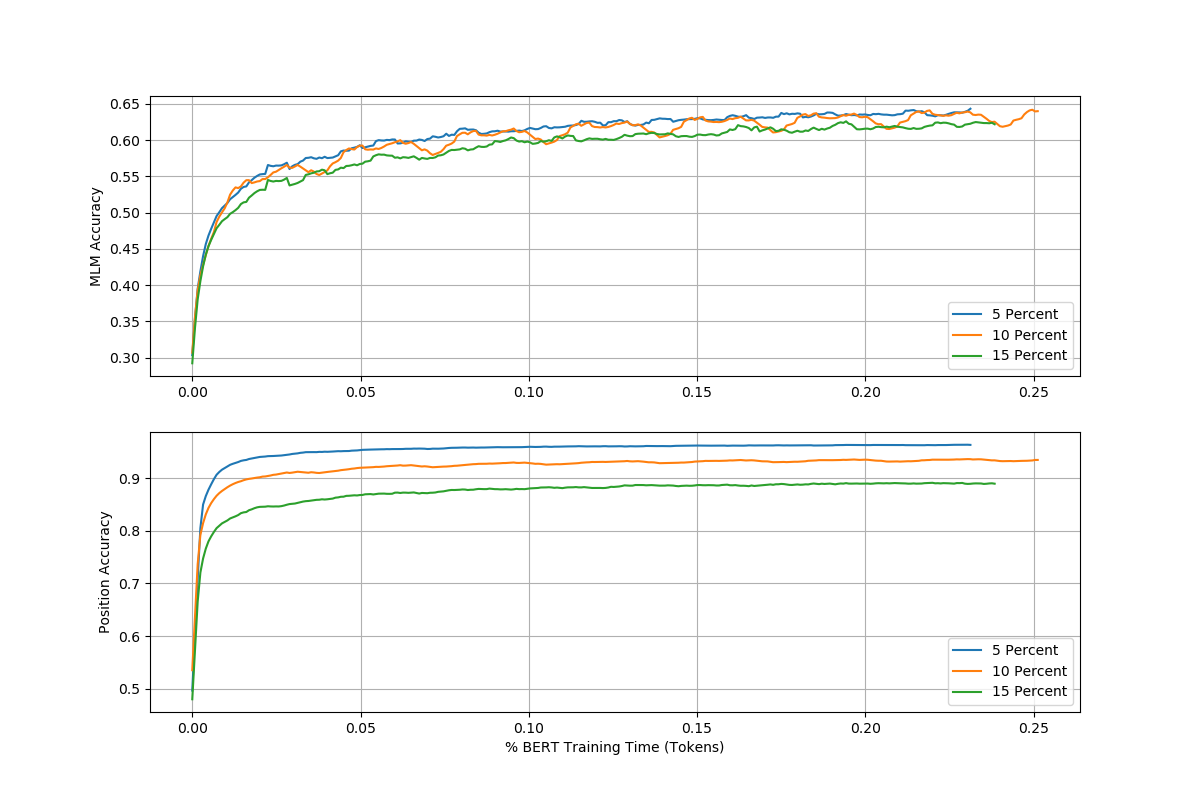}
    \caption{IPU percentage sweep} \label{fig3:side:a}
    \end{minipage}%
    \hspace{1cm}%
    \begin{minipage}[t]{0.4\linewidth}
    \centering
    \includegraphics[width=1.0\columnwidth]{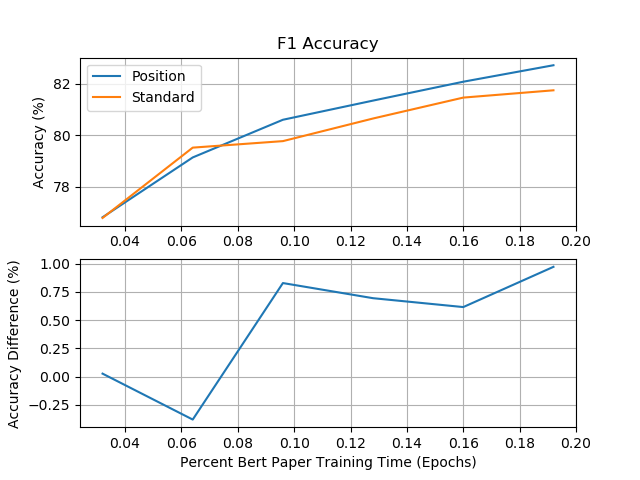}
    \caption{Phase 1 position improvement} \label{fig3:side:b}
    \end{minipage}
  \rule[-.5cm]{4cm}{0cm}
\end{figure}

\begin{figure}
  \centering
  \rule[-.5cm]{0cm}{4cm} 
   \begin{minipage}[t]{0.4\linewidth}
    \centering
    \includegraphics[width=1.0\columnwidth]{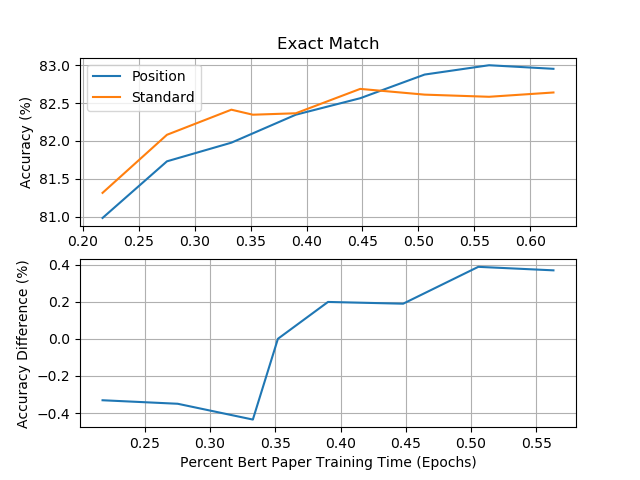}
    \caption{Phase 2 exact match improvement} \label{fig4:side:a}
    \end{minipage}%
    \hspace{1cm}%
    \begin{minipage}[t]{0.4\linewidth}
    \centering
    \includegraphics[width=1.0\columnwidth]{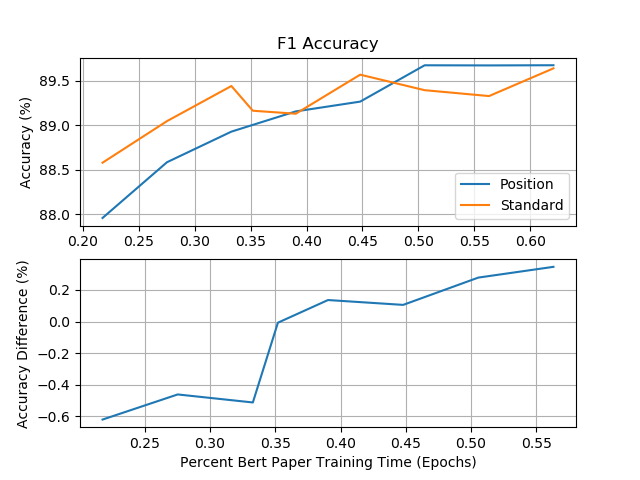}
    \caption{Phase 2 F1 score improvement} \label{fig4:side:b}
    \end{minipage}
  \rule[-.5cm]{4cm}{0cm}
\end{figure}

The results for phase 1 of the Squad downstream task are shown in figure \ref{fig3:side:b}. There is approximately a 1\% performance advantage at about 20\% of the original training time for BERT Base. Figure \ref{fig4:side:a} and figure \ref{fig4:side:b} shows the results of phase 2 training for this run which converge to a ~.3\% performance advantage matching the GPU simulations. The convergence time is 10\% faster with position masking which is likely caused by the tighter control over the position encodings.

\subsection{Position Mask Percentage Comparison}

\begin{figure}
  \centering
  \rule[-.5cm]{0cm}{4cm} 
    \begin{minipage}[t]{0.4\linewidth}
    \centering
    \includegraphics[width=1.0\columnwidth]{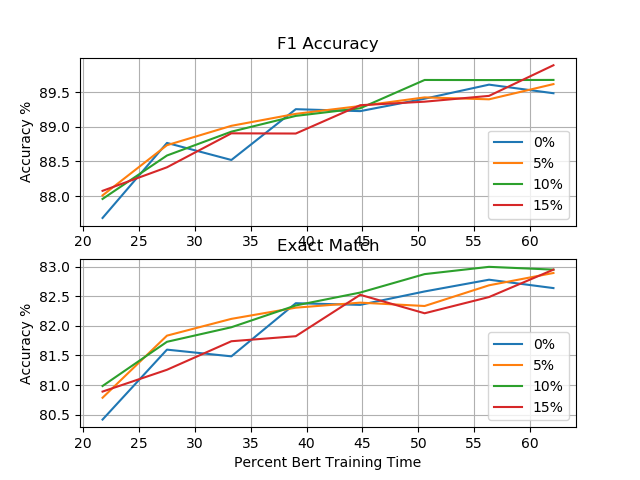}
    \caption{Convergence time} \label{fig5:side:a}
    \end{minipage}%
    \hspace{1cm}%
    \begin{minipage}[t]{0.4\linewidth}
    \centering
    \includegraphics[width=1.0\columnwidth]{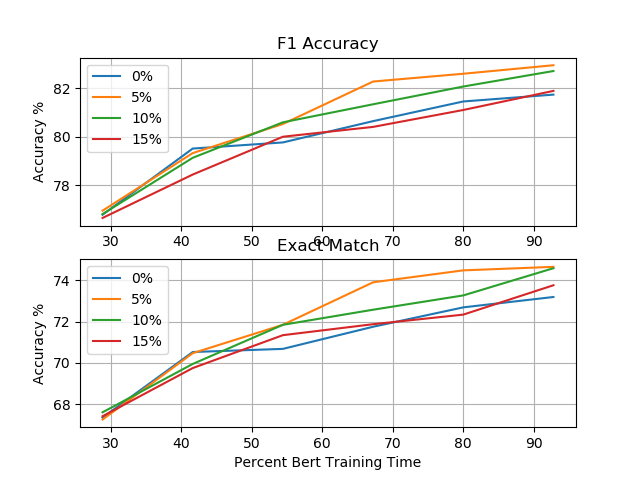}
    \caption{Phase 1 masking sweep} \label{fig5:side:b}
    \end{minipage}
  \rule[-.5cm]{4cm}{0cm}
\end{figure}

Figure \ref{fig3:side:a} shows a comparison of the MLM and position accuracy when the position masking percentage was varied from 5\% to 15\%. As expected the greater masking leads to a lower MLM and position accuracy. The majority of the work was done using 10\% masking percentage which led to the best results. Figures \ref{fig5:side:a} and \ref{fig5:side:b} show a result from a sweep over the position masking percentage. For phase 1 a 10\% masking percentage led to the best results with a 15\% masking percentage leading to equivalent performance as no masking. The 5\% masking showed similar results to the 10\% masking results.  

\subsection{Enhanced Dropout Gradient Approach}

\begin{figure}
  \centering
  \rule[-.5cm]{0cm}{4cm} 
    \begin{minipage}[t]{0.4\linewidth}
    \centering
    \includegraphics[width=1.0\columnwidth]{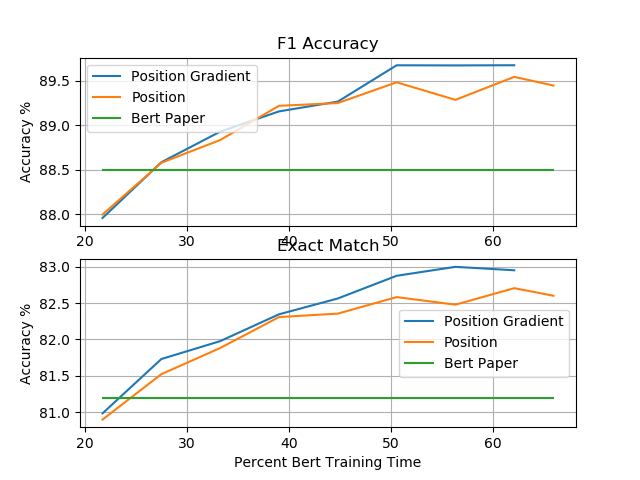}
    \caption{Default Dropout Gradient} \label{fig7:side:a}
    \end{minipage}%
    \hspace{1cm}%
    \begin{minipage}[t]{0.4\linewidth}
    \centering
    \includegraphics[width=1.0\columnwidth]{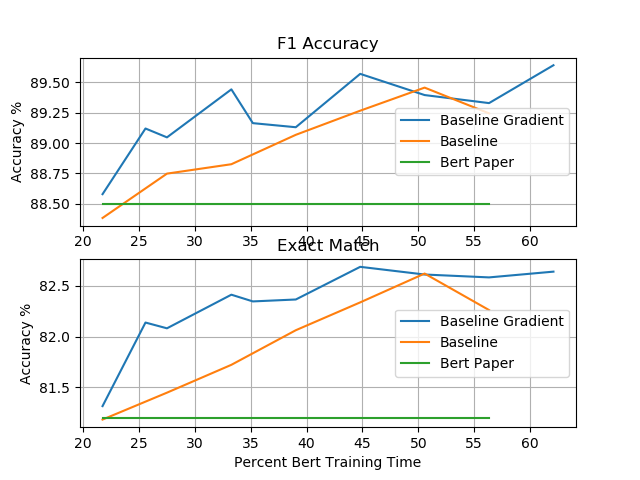}
    \caption{Position Dropout Gradient} \label{fig7:side:b}
    \end{minipage}
  \rule[-.5cm]{4cm}{0cm}
\end{figure}

Figure \ref{fig7:side:a} shows a comparison for results with and without the enhanced dropout gradient approach for the default BERT Base with Figure \ref{fig7:side:b} showing the results with position masking.We have analysed the gradients at the output of the Softmax operation starting from the same pre-trained weights, and find them to be higher for the case when the dropout is not applied. We are unsure whether this give some sort of a regularisation effect due to the small size of the data sets. The reasons for the gains need further study and exploration as the gains are not insignificant.

 \begin{figure}
  \centering
  \rule[-.5cm]{0cm}{4cm} 
   
    \begin{minipage}[t]{0.4\linewidth}
    \centering
    \includegraphics[width=1.0\columnwidth]{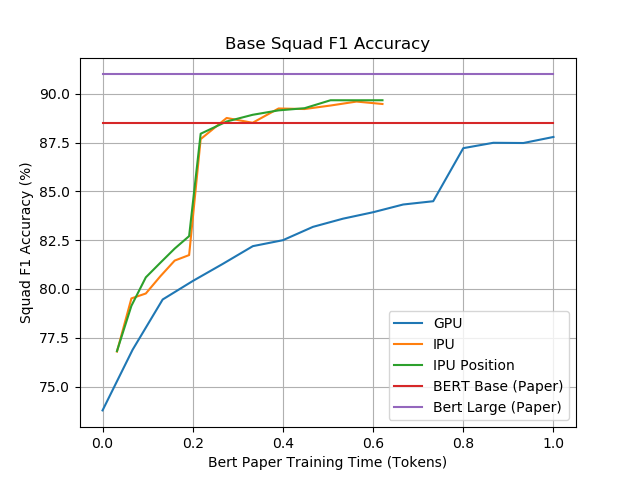}
    \caption{IPU vs GPU Performance} \label{fig11:side:a}
    \end{minipage}%
    \hspace{1cm}%
  \rule[-.5cm]{4cm}{0cm}
\end{figure}

\subsection{IPU vs GPU Performance}

The IPU implementation has a convergence and steady state advantage over the GPU for the BERT BASE implementation with both position masking and in general. Figure \ref{fig11:side:a} shows a comparison between the IPU Squad performance, GPU performance and results from the original BERT paper. The IPU has a convergence time advantage as well as a steady state performance advantage. The reasons for the gains are not currently clear and require further study.

\section{Conclusion}

We have shown that masking positions as well as tokens leads to both a convergence time advantage as well as steady state accuracy improvement. In the future we plan to map this to other architectures to determine if the performance advantage scales. We also feel that masking the position opens up a new dimension for optimizing transformer networks that hopefully can open up new ideas which greatly improve performance.

\section*{Broader Impact}

This paper is an algorithmic improvement upon an existing architecture and doesn't have a broad impact other than enhancing existing techniques.

\bibliographystyle{unsrt}  
\bibliography{references}
\end{document}